# A novel MDPSO-SVR hybrid model for feature selection in electricity consumption forecasting

Yukun Bao[1], Liang Shen[1], Xiaoyuan Zhang[2*], Yanmei Huang[3], Changrui Deng[3]

[1]Center for Modern Information Management,
School of Management, Huazhong University of Science and Technology, Wuhan, P.R.China, 430074

[2] Department of Public Curriculum Studies, Jiangxi University of Software Professional Technology, Nanchang 330041, P.R. China

[3] Center of Big Data Analytics, Jiangxi University of Engineering, Xinyu 338029, P.R. China

**Abstract**

Electricity consumption forecasting has vital importance for the energy planning of a country. Of the enabling machine learning models, support vector regression (SVR) has been widely used to set up forecasting models due to its superior generalization for unseen data. However, one key procedure for the predictive modeling is feature selection, which might hurt the prediction accuracy if improper features were selected. In this regard, a modified discrete particle swarm optimization (MDPSO) was employed for feature selection in this study, and then MDPSO-SVR hybrid model was built to predict future electricity consumption. Compared with other well-established counterparts, MDPSO-SVR model consistently performs best in two real-world electricity consumption datasets, which indicates that MDPSO for feature selection can improve the prediction accuracy and the SVR equipped with the MDPSO can be a promised alternative for electricity consumption forecasting.

* Corresponding author: Tel: +86-15170485489.
Email: xiaoyuanzhang1224 @163.com

# 1. Introduction

Electricity consumption forecasting is of vital importance for the energy planning of a country since it can effectively facilitate the balance of the generation and supply of electric energy in a long run. As such, it has been considered as a very important issue, and various predictive models have been proposed during the past few decades, such as regression models [1, 2], exponential smoothing models [3-5], the auto-regressive moving average models [6], fuzzy logic [7], grey models [8, 9], artificial neural networks (ANN) [10, 11], and support vector machines (SVM) [12-15]. Among these, support vector regression (SVR) is a powerful machine learning technique with strong theoretical foundation [16] and is employed as the modeler for electricity consumption forecasting in this study.

One of the major procedures of predictive modeling with SVR is the selection of input features, which is important for the prediction accuracy and thus motivates this present research. In the existing literature, feature selection techniques can be divided into two major types on the basis of the evaluation criterion of selecting feature subset: filter methods and wrapper methods. The commonly used evaluation criterions in filter methods contain mutual information [17], Bayesian automatic relevance determination [18], and correlation and linear independency [19], which mainly determine a set of input features based on the correlation between each subset of input variables and the output. Here, filter methods do not be involve with model learning guided by prediction accuracy. Due to this, a data-driven forecasting model with filter methods often fails to generate acceptable forecasts. As opposed to the filter methods, the wrapper methods take the performance of the forecasting model as an evaluation criterion to identify the correct input subset from a big pool of candidate features [20].

Obviously, the implementation of wrapper methods will result in a vast search space that is well suited for stochastic optimization techniques.

Particle swarm optimization algorithm (PSO), as a popular stochastic optimization technique, has motivated a variety of studies in the filed of optimization problems, owning to its simple structure [21, 22], easy of implementation, and can produce "acceptable good" solutions at a very low computational cost [23, 24]. Earlier studies on PSO are mainly concentrated on those problems with pure continuous variables, however, decision variables of some problems are binary in nature and just can be represented by binary coding, such as the selection of input features. In this regard, Kennedy first designed a strategy to operate on discrete binary variables based on the original PSO, which is called discrete particle swarm optimization (DPSO) [25]. This approach has been widely applied in discrete power system problems [26-28] and implemented for features selection [29-31] due to its simple structure. It should be noted that there are two demerits in the DPSO. One is that the nonlinear sigmoid function in the algorithm is much sensitive to high velocities, and the other is that the trajectory of one-dimensional particle in the algorithm is in a state of convergence and divergence (i.e., when a particle reaches the near-optimal solution, it will disperse more quickly).

Unlike DPSO, Afshinmanesh et al. [32] first defined velocities of particles as the changes in bits of a binary string by introducing 'or' and 'xor' operators of Boolean algebra, and the approach is called as Boolean PSO, which does not incorporate any nonlinear function and is shown to outperform the DPSO and genetic algorithm (GA) in some well-known test functions. Afterwards, Marandi et al. [33] added the notion of inertia weight on the flight equations of the above Boolean PSO by introducing 'and' operator of Boolean algebra and applied this strategy to a hard multi-objective electromagnetic problem. To overcome the second demerit of the DPSO, Xu et al [34] proposed a modified discrete particle swarm optimization (MDPSO) by redefining the position update equation of particles, and verified that when a particle reaches the near-optimal solution, its divergence is relatively weak. Besides, there exist some other approaches for tacking discrete binary variables, such as Angle-modulated PSO [33], Quantum PSO [35], Binary PSO based on estimation of distribution [36] and so on, more related information can be referred by [37]. Following the research line of MDPSO, in this study, we proposed a novel MDPSO-SVR hybrid model for feature selection to forecast electricity consumption.

The contributions of this study can be summarized as follows. a) We proposed a novel MDPSO-SVR hybrid model for feature selection. While there are numerous studies on DPSO and SVR respectively, limited work, if any, has investigated the hybrid of MDPSO and SVR for feature selection in the literature concerning electricity consumption forecasting. b) We design four different feature selection strategies for comparative analysis to confirm the benefit of MDPSO in SVR modeling. c) Compared with other well-established counterparts, MDPSO-SVR consistently performs best in two real-world electricity consumption datasets, which indicates that MDPSO for feature selection can improve the prediction accuracy of SVR modeling and the SVR equipped with the MDPSO can be a promising alternative for electricity consumption forecasting.

This paper is structured as follows. In section 2, SVR model, DPSO and MDPSO are introduced in the subsections 2.1, 2.2 and 2.3 respectively. Section 3 elaborates on the proposed MDPSO-SVR hybrid modeling framework. Section 4 shows the details of experimental datasets, candidate features, performance measures, experimental design and experimental implementation. Following that, the results and discussion are reported in Section 5. Finally, we conclude the paper in section 6.

## 2. Methodologies

In this section, SVR model, DPSO and MDPSO are introduced in the subsections 2.1, 2.2 and 2.3 respectively.

### 2.1 Modeling with Support Vector Regression (SVR)

The essence of electricity consumption forecasting is a type of regression process. Consider the input sample set $\{(x_i, y_i)\}_{i=1}^{n}$, where $x_i \in R^m$ is the $i$-th vector containing $m$ input features, $y_i$ is the corresponding desired consumption, and $n$ denotes the number of data items in the sample set. SVR modeling firstly map the input sample $x$ to a high-dimensional feature space by nonlinear mapping function $\varphi(x)$, and then build a linear model in the feature space to estimate the regression function. The formula is as follows.

$$f(x,w) = w \cdot \phi(x) + b \qquad (1)$$

where $w$ is weight vector, $b$ is threshold value. For the given training data set $\{(x_i, y_i)\}_{i=1}^{n}$, SVR with $\varepsilon$ insensitive loss function is called $\varepsilon - SVR$, and its constraint optimization problem is as follows.

$$\min \quad R = \frac{1}{2}\|w\|^2 + C\sum_{i=1}^{n}(\xi_i + \xi_i^*)$$

$$\text{s.t.} \quad \begin{cases} (w\phi(x_i) + b) - y_i \leq \varepsilon + \xi_i^* \\ y_i - (w\phi(x_i) + b) \leq \varepsilon + \xi_i \\ \xi_i, \xi_i^* \geq 0, \ i = 1, 2, \ldots, n \end{cases} \qquad (2)$$

where $C$ is a penalty parameter, whose value determines the tolerance of outliers. Here $\xi_i$

and $\xi_i^*$ are non-negative slack variables, and $\varepsilon$ is the deviation between the regression function $f(x_i)$ and the actually obtained targets $y_i$ for all the training data.

The optimization problem of formula (2) can be transformed into a dual problem by introducing Lagrangian function. By solving the dual problem, we can get the solution of formula (1).

$$f(x) = \sum_{i=1}^{n_{SV}} (\alpha_i - \alpha_i^*) K(x_i, x) + b \qquad (3)$$

where $\alpha_i, \alpha_i^*$ are non-negative Lagrange multipliers. Only a small part of $\alpha_i, \alpha_i^*$ is not zero, and their corresponding sample is the support vector. $n_{SV}$ is the number of support vector. Here $K(x_i, x)$ is defined as the kernel function, whose value equals the inner of the vectors $x_i, x$ in the feature space $\phi(x_i), \phi(x)$. Radial basis function is usually used in all kernel functions.

$$K(x_i, x) = \exp(-\gamma \|x_i - x\|^2) \qquad (4)$$

where $\gamma$ is the kernel parameter.

## 2.2 Discrete Particle Swarm Optimization algorithm (DPSO)

In the DPSO, each particle in the population contains two attributes, position and velocity, where the position of particle is constrained to a state space that can only take 0 and 1, and the velocity of particle is a change probability of its position. Each particle adjusts its flight velocity using the following update equations (5). Afterwards, the velocity of particle is constrained to the interval [0,1] by using the following sigmoid transformation:

$$V_{id}^{t+1} = w \cdot V_{id}^t + c_1 \cdot r_{1d} \cdot (pbest_{id}^t - P_{id}^t) + c_2 \cdot r_{2d} \cdot (gbest_d^t - P_{id}^t) \qquad (5)$$

$$S(V_{id}^{t+1}) = \frac{1}{1 - \exp(-V_{id}^{t+1})} \qquad (6)$$

where $V_{id}^t$ and $P_{id}^t$ represent the velocity and the position of the $d$th dimension of the $i$th particle in the $t$th iteration respectively. $pbest_{id}^t$ and $gbest_d^t$ denotes the optimal position visited by the particle $i$ and the whole swarm up to the $t$th generation respectively. $c_1$ and $c_2$ are the acceleration constants which reflect the weighting of particles' learning ability from cognitive and interaction information. $r_{1d}$ and $r_{2d}$ are two random real numbers in the range [0, 1].

The DPSO updates the position of particle according to the equation (7), as follows:

$$P_{id}^{t+1} = \begin{cases} 1 & if\ rand < S(V_{id}^{t+1}) \\ 0 & otherwise \end{cases} \quad (7)$$

where $rand$ is a random real number uniformly distributed between 0 and 1. $S(V_{id}^{t+1})$ represents the probability of bit $P_{id}^{t+1}$ taking one. To avoid $S(V_{id}^{t+1})$ approaching 0 or 1, the constant $V_{max}$ is retained in the DPSO, that is $|V_{id}^{t+1}| < V_{max}$, which controls the ultimate mutation rate of bit vector. For instance, when $V_{max}$ is equal to 0.2, the value range of $S(V_{id}^{t+1})$ will be between 0.4502 and 0.5498, which indicates that the probability of the bit $P_{id}$ taking 0 and 1 is approximately equal, which will cause position state of the particle to be in an irregularly divergent state, which is not conducive to population convergence. When $V_{max}$ is equal to 6, $S(V_{id}^{t+1})$ will be limited to the interval [0.0025, 0.9975], which means that as $V_{id}^{t+1}$ gets larger, $S(V_{id}^{t+1})$ will approach 1 (i.e., the probability of bit $P_{id}^{t+1}$ taking one is close to 1), on the contrary, as $V_{id}^{t+1}$ gets smaller, $S(V_{id}^{t+1})$ will approach 0 (i.e, the probability of bit $P_{id}^{t+1}$ taking one is close to zero), which indicates the DPSO is much sensitive to high velocities. For this, $V_{max}$ is usually set at 4 [38].

Next, the divergence of the DPSO algorithm will be further analyzed. Since each dimension of particles is independent in the optimization process, for the convenience of description, supposing that the number of dimensions of particles is 1, the dimension subscript of particles is removed, and the equation (5) is processed as follows:

$$V^{t+1} = w.V^t + c_1.r_1.(pbest^t - P^t) + c_2.r_2.(gbest^t - P^t)$$

Let $\varphi_1 = c_1.r_1$, $\varphi_2 = c_2.r_2$, $\delta = \varphi_1.(pbest^t - P^t) + \varphi_2.(gbest^t - P^t)$, then $V^{t+1} = w.V^t + \delta$

Let $V^t = V_0$, then $V^{t+1} = w.V_0 + \delta$, $V^{t+2} = w^2.V_0 + w.\delta + \delta$, ..., $V^{t+k} = w^k.V_0 + \sum_{i=1}^{k} w^{i-1}.\delta$

($k \geq 1$). When the particle is in the best position of current search, i.e., $P^t = pbest^t = gbest^t$, obviously $\delta = 0$ and $V^{t+k} = w^k.V_0$. Since $0 < w < 1$, according to the limit theory, when $k$ reaches a certain constant $M > 0$, there is a very small positive constant $\varepsilon$, such that $w^k < \varepsilon$ (namely $w^k \approx 0$), so $V^{t+k} \approx 0$. At this time, it can be seen from the equation (6) that $S(V^{t+k})$ will approach 0.5. According to the formula (7), it can be known that the probability of the bit $P^{t+k}$ taking zero and one is approximately equal, which means that when the particle in the DPSO reaches the optimal solution, it will disperse more quickly, which is not conducive to the local search of the population in the later stage of evolution. To solve this inherent problem in the DPSO, MDPSO was proposed by Xu et al [34], which will be introduced in the following subsection 2.3.

## 2.3 Modified Discrete Particle Swarm Optimization algorithm (MDPSO)

On the basis of the DPSO, the MDPSO introduces two new operations: position operation $X \oplus Y = \begin{cases} 1 & \text{if } X=Y \\ -1 & \text{else} \end{cases}$ and position transfer $-X = \begin{cases} 0 & \text{if } X=1 \\ 1 & \text{if } X=0 \end{cases}$, where X and Y are the position values of any two particles in the same dimension. In the MDPSO, each particle adjusts its flight velocity using the following update equations (8) and (9).

$$V_{id}^{t+1} = w.V_{id}^t + c_1.r_{1d}.(pbest_{id}^t \oplus P_{id}^t) + c_2.r_{2d}.(gbest_d^t \oplus P_{id}^t) \qquad (8)$$

$$P_{id}^{t+1} = \begin{cases} P_{id}^t & \text{if } rand < S(V_{id}^t) \\ -P_{id}^t & \text{otherwise} \end{cases} \tag{9}$$

where the meanings of symbols in formulas (8) and (9) are consistent with the DPSO, which are omitted here to save space.

The motion state analysis of particles in the DPSO is also applied to the MDPSO. Let $\varphi_1 = c_1 \cdot r_1$, $\varphi_2 = c_2 \cdot r_2$, $\delta' = \varphi_1 \cdot (pbest^t \oplus P^t) + \varphi_2 \cdot (gbest^t \oplus P^t)$, then $V^{t+1} = w \cdot V^t + \delta'$. When a particle is in the current optimal position, i.e., $P^t = pbest^t = gbest^t$, $\delta' = \varphi_1 + \varphi_2$, the update speed $V^{t+1}$ of the particle will be the largest, and the particle will maintain the original state with a large probability $S(V^{t+1})$, which is conducive to the local search of the population. On the other hand, the particle will also mutate with probability $1 - S(V^{t+1})$ to reach a new state in which $pbest^t = gbest^t \neq P^t$, $\delta' = -(\varphi_1 + \varphi_2)$. At this time, the update velocity $V^{t+1}$ of the particle will be the smallest. According to formula (9), it can be known that the particle will perform state transition with a large probability of $1 - S(V^{t+1})$, showing a trend of convergence to the optimal value, which is a kind of purposeful optimization process.

## 3. The proposed MDPSO-SVR hybrid modeling framework

The MDPSO-SVR hybrid modeling framework was developed in this study. MDPSO takes the prediction performance of SVR as an evaluation criterion to identify the correct input subset from a big pool of candidate features. For clarity, the flowchart of the overall learning process of the MDPSO-SVR modeling framework is depicted in the Fig.1.

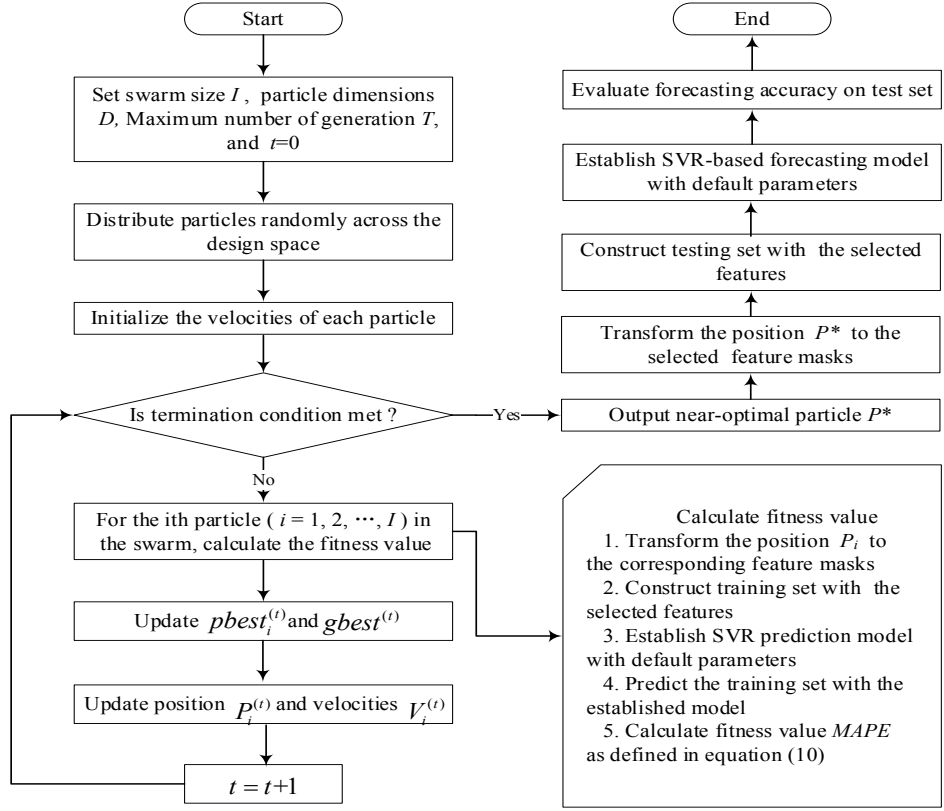

Fig. 1. Flowchart of the MDPSO-SVR hybrid modeling framework

In the following subsections, the details of implementation of the proposed MDPSO-SVR strategy for electricity consumption forecasting are described.

### 3.1 Initialization

In the MDPSO, the position and velocity of each particle can be described as $P_i = (P_{i,1}, P_{i,2},...,P_{i,n_f})$ and $V_i = (V_{i,1}, V_{i,1},...,V_{i,n_f})$, where $P_{i,1},...,P_{i,n_f}$ represents the feature masks of the $i$th particle and the value of 1 or 0 for each feature mask denotes the corresponding feature is selected or excluded. Here $n_f$ is the number of bits denoting the number of feature candidates. Initially, the particles in the population are randomly distributed across a designed $n_f$ dimensional space and each dimension corresponds to a specified feature that is randomly generated with 0 or 1 using a 50% probability, respectively.

The initial velocities of all particles are randomly assigned a real value which equals to the product of a random number and the maximum velocity $V_{max}$, here, $V_{max}$ is set at 4 to avoid $S(V_{id})$ approaching 0 or 1.

### 3.2 Evaluation

As we all know that the objective of the SVR forecasting model is to predict future electricity consumption with high accuracy. Thus, it is critical to choose a fitness function that can estimate the generalization ability when determining the model's input features using the proposed MDPSO-SVR strategy. Considering that the mean absolute percentage error (MAPE) index can efficiently measure the performance of prediction model in a scale-free way, in this study, we use MAPE as the fitness function of the MDPSO-SVR strategy.

In the MDPSO-SVR strategy, take the $i$th particle as an example, it's position $P_i = (P_{i,1}, P_{i,2}, ..., P_{i,n_f})$ is firstly transformed to the corresponding feature subset. Then the selected features are used to construct the input variables of training set and the default parameters are used to establish SVR forecasting model. Once all prediction values in training set are obtained, the fitness value of the $i$th particle is calculated according to $MAPE = \frac{1}{N}\sum_{t=1}^{N}\left|\frac{y_t - \hat{y}_t}{y_t}\right| \times 100\%$, where $N$ is the length of training set, $y_t$ is the actual value at period $t$, $\hat{y}_t$ is the prediction value at period $t$.

### 3.3 Update

In the current generation, once the fitness values of all the particles in the population are evaluated, it is not difficult to get the *pbest* of each particle and the *gbest* of the swarm according to the definition of *pbest* and *gbest*. Then the velocity and position of each particle are updated according to the equations (8) and (9), respectively. Afterwards, the fitness values of all the particles are recalculated as per the subsection 3.2, and the *pbest* of each particle and the *gbest* of the swarm are updated in turn. A judgement is made as to whether the termination condition is satisfied and, if so, the optimal particle with the correct feature set is returned; otherwise, the previous step is repeated.

**3.4　Validation**

Once the optimal particle with the correct feature set is obtained, the forecasting values on test set will be evaluated with it. Specially, the optimal particle' position is firstly transformed to the corresponding features. Then the selected features are used to construct the input variables of testing set and the default parameters are used to establish SVR forecasting model. Next, the output variables of testing set will be predicted with the above input variables and forecasting model. When all prediction values are obtained, three alternative forecasting accuracy measures in the following subsection 4.3 will be calculated to further assess the forecasting performance of the model.

**4. Experimental design and setup**

**4.1　Data description**

To verify the forecasting performance of the proposed MDPSO-SVR strategy, two real-world datasets, i,e., monthly electricity consumption data from United States and China, were used in this present study. The period of the first data set ranges from Jan. 2005 to Sep. 2019 which is divided into the training set (Jan. 2005 to Dec. 2016) and testing set (the last 33 months), and that of the second data set is from Jan. 2010 to Dec. 2019 which is divided into the training set (Jan. 2010 to Jul. 2018) and testing set (the last 17 months). To simulate real-world scenarios, the test set was considered to be completely independent to the training sets and didn't get involved in the learning procedure. The first data can be obtained from the website of the United States Energy Administration (http://www.eia.doe.gov) and is presented in Fig. 2(a). And the second data is National Energy Administration (http://www.nea.gov.cn), which is depicted in Fig. 2(b).

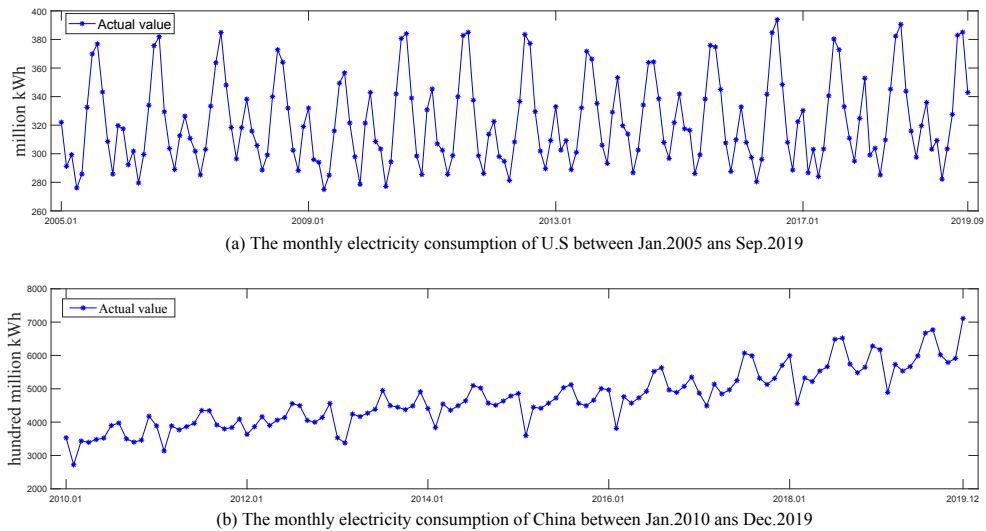

(a) The monthly electricity consumption of U.S between Jan.2005 ans Sep.2019

(b) The monthly electricity consumption of China between Jan.2010 ans Dec.2019

Fig. 2 The monthly electricity consumption of U.S and China

### 4.2 Candidate features

Considering the short-run trend and yearly periodicity characteristics of monthly electricity consumption, values for the previous 12 months and similar months in the previous years were selected as the input variables set for the forecasting model. Consequently, the candidate set of input variables can be summarized as follows:

$$Input(t) = \{y(t-1), y(t-2), ..., y(t-12), y(t-24), y(t-36)\}$$

Where $y(t-i)$ denotes the lagged electricity consumption of $t-i$. Obviously, there are 14 candidate inputs.

### 4.3 Performance measure

To assess the forecasting performance of a model, three alternative forecasting accuracy measures were employed in this study, which are the Mean Absolute Percentage Error (MAPE), the Root Mean Squared Error (RMSE), and the Normalised Root Mean Squared Error (NRMSE)[39] respectively. Definitions of these measures are presented in the following expressions (10)-(12).

$$MAPE = \frac{1}{N}\sum_{t=1}^{N}\left|\frac{y_t - \hat{y}_t}{y_t}\right| \times 100\% \qquad (10)$$

$$RMSE = \sqrt{\frac{1}{N}\sum_{t=1}^{N}(y_t - \hat{y}_t)^2} \qquad (11)$$

$$NRMSE = \sqrt{\frac{1}{N}\sum_{t=1}^{N}(y_t - \hat{y}_t)^2} \Big/ \bar{y} \qquad (12)$$

Where $y_t$ and $\hat{y}_t$ are the actual value and the prediction value at period $t$ respectively. The smaller the three measure values are, the closer are the predicted values to the actual values.

**4.4  Experimental design**

There are two goals in the experimental study. One is to confirm the benefits of the MDPSO for feature selection in SVR modeling. To achieve the aim, SVR as the modeler, the commonly used conventional DPSO, boolean PSO (BPSO) and genetic algorithm (GA) are used for comparative analysis with the MDPSO. The other is to examine the forecasting performance of the proposed MDPSO-SVR strategy, for this, Holt-Winters additive model from time series forecasting models and back propagation neural network model (BPNN) based on machine learning techniques are selected as well-established counterparts to further compare with the proposed method. Here, the smoothing parameters of the Holt-Winters additive model are optimized by continuous-valued PSO. For convenience, all the designed

models were abbreviated as follows:

(1) SVR: the SVR model with a full input vector and the default parameters values, which means there is no feature selection.

(2) DPSO-SVR: the SVR model combined feature selection based on the conventional DPSO with the default parameters.

(3) BPSO-SVR: the SVR model combined feature selection based on the Boolean PSO with the default parameters.

(4) GA-SVR: the SVR model with GA based feature selection and the default parameters.

(5) MDPSO-SVR: the SVR model combined feature selection based on the MDPSO with the default parameters.

(6) Holt-Winter: the Holt-Winters additive model with PSO based parameter optimization.

(7) BPNN: the BPNN model with a full input vector and the default parameters values.

(8) MDPSO-BPNN: the BPNN model combined feature selection based on MDPSO with the default parameters.

In the above approaches, model (1), as the benchmark, generates the forecast without feature selection. Compared model (1), models (2-5) employ different strategies for feature selection respectively, where models (2-4) are used for comparative analysis to further verify the superiority of the MDPSO. Models (6-8), as the well-established counterparts, are designed to examine the forecasting performance of the proposed MDPSO-SVR strategy.

**4.5 Experimental implementation**

All our experiments were carried out in MATLAB 2016a and executed on a computer with Intel Core i5-3230M, 2.60 GHz CPU and 4GB RAM. Among the above eight models, LibSVM (version 2.86) was employed for SVR modeling, and nnet toolbox that comes with MATLAB 2016a was used for BPNN modeling. Additionally, the matlab codes of DPSO, BPSO, GA, MDPSO and Holt-Winters methods were written in source code. According to a trial-error fashion considering the trade-off between prediction accuracy and computational time, the parameters in PSO and its variants are set as follows: the population size is set to 20; the number of total iterations is set to 200; The acceleration coefficients c1 and c2 are both equals to 2; the inertial weight varied linearly from 0.9 to 0.4 according to the generations. The parameters in GA are set as follows: the population size is also set to 20; the number of total iterations is set to 200; the probability of crossover and mutation is set to 0.90 and 0.09 respectively. It should be noted that the code (and data) in this study has been certified as Reproducible by Code Ocean: (https://codeocean.com/). Permanent link to reproducible Capsule can be retrieved with https://doi.org/10.24433/CO.9592075.v1., and the readers can replicate this study on their own.

Considering that the PSO and its variants belong to random search algorithms, that is, the result of each run is different because of the different random number, which leads to the instability of the optimization result, the modeling process for each method is repeated 50 times, all examined models are tested with the hold-out samples, and their prediction performance is judged by the mean and standard deviationof each accuracy measure.

## 5. Results and discussion

*5.1 Study 1: Examination of the MDPSO for feature selection in SVR model*

To confirm the superiority of the MDPSO, in this paper, the commonly used DPSO,

BPSO and GA are used for comparative analysis, and the quantitative and comprehensive assessments are performed with real-world electricity consumption dataset based on prediction accuracy and computational cost. Table 1 describes the forecasting performance of the pure SVR model without feature selection method and the SVR models with the above four different feature selection methods on U.S and China datasets, which is measured by the means and standard deviations of three accuracy indexes (i.e., MAPE, RMSE, NRMSE) for 50 times, here, Full input represents the pure SVR model with a full input vector and the default parameters values, and the last column shows the elapsed time of all the examined models. Additionally, taking the accuracy index MAPE as an example, this paper intuitively presents the prediction performance of four examined models on the out-of-sample in each modeling process, as shown in Fig. 3(a-d) and Fig. 4(a-d). Here, the red asterisk and red circle mark the minimum and maximum values of MAPE sequence respectively.

**Table 1.** Prediction accuracy measures of all the examined models in U.S and China datasets

| Dataset | Strategy | Accuracy | | | Elapsed Time |
|---|---|---|---|---|---|
| | | MAPE | RMSE | NRMSE | |
| U.S | Full input | 1.92 | 8.59 | 0.0263 | **0.11** |
| | DPSO | 1.99±0.14 | 8.83±0.52 | 0.0270±0.0016 | 0.98 |
| | BPSO | 1.95±0.14 | 8.73±0.49 | 0.0267±0.0015 | 1.09 |
| | GA | 1.95±0.12 | 8.71±0.39 | 0.0267±0.0012 | 545.31 |
| | MDPSO | **1.82±0.04** | **8.28±0.22** | **0.0253±0.0006** | 2.01 |
| China | Full input | 2.85 | 256.86 | 0.0428 | **0.13** |
| | DPSO | 2.78±**0.004** | **237.97±2.07** | **0.0396±0.0003** | 2.22 |
| | BPSO | 2.79±0.068 | 244.47±7.08 | 0.0407±0.0012 | 2.19 |
| | GA | 2.90±0.201 | 252.35±13.7 | 0.0421±0.0023 | 149.25 |
| | MDPSO | **2.77**±0.020 | 242.54±6.45 | 0.0404±0.0011 | 4.43 |

Note: For each row of table, the entry with smallest value is set in boldface.

According to the results presented in Table 1, we can reduce the following observation: In U.S case, when considering the means of the three accuracies, MDPSO consistently is superior to Full input, Full input consistently is better than GA and BPSO, here, GA and BPSO is almost a tie, and DPSO performs worst. Additionally, it is obvious that the ranking with respect to standard deviations of the three accuracies is MDPSO, then GA, then BPSO, and then DPSO. In China case, DPSO, MDPSO and BPSO consistently outperform Full input in term of the means of the three accuracies. As for as the comprison of DPSO, MDPSO, BPSO and GA, the ranking in term of mean and standard deviation of the three accuracies both are DPSO, then MDPSO, then BPSO, and then GA. With respect to the elapsed time, there is same conclusion in the above two datasets, that is, GA is computationally much more expensive than the other three feature selection strategies, MDPSO is twice as fast as DPSO, DPSO and BPSO are almost a tie, and Full input has the least computation time.

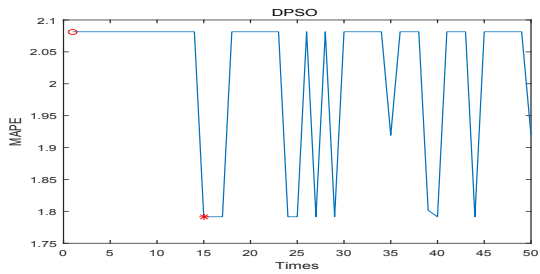

Fig. 3(a) MAPE comparison of DPSO

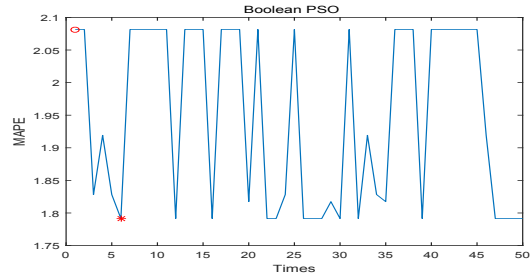

Fig. 3(b) MAPE comparison of Boolean PSO

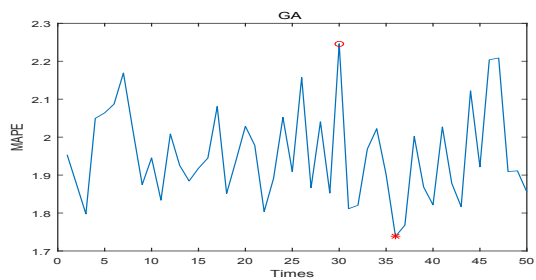

Fig. 3(c) MAPE comparison of GA

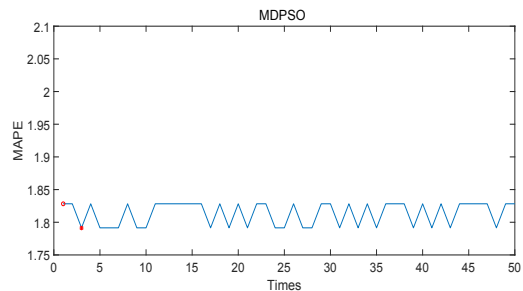

Fig. 3(d) MAPE comparison of MDPSO

Fig. 3 MAPE comparison of four examined models in each modeling processing in U.S dataset

As we can see from the Fig. 3(a-b), there are sharp oscillations in DPSO and BPSO, with a maximum of 2.081 and a minimum of 1.791. According to the Fig. 3(c), it is clear that GA fluctuates violently between 2.246 and 1.738, and the MAPE value of GA is different in each experiment. Unlike GA, MDPSO in Fig. 3(d) has only two results, 1.828 and 1.791, in 50 modeling processes, and it fluctuates between the two values. Here, a small MAPE value means that DPSO, BPSO, GA or MDPSO selects a correct set of features, and vice versa. According to MAPE value, it is obvious that DPSO, BPSO and GA are inferior to Full input in the majority of experiments, while MDPSO consistently outperforms Full input in all experiments in U.S dataset. In addition, by comparing the fluctuation ranges of DPSO, BPSO, GA and MDPSO, it is not difficult to find that MDPSO has the smallest fluctuation range.

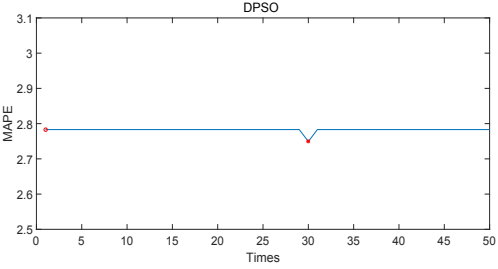

Fig. 4(a) MAPE comparison of DPSO

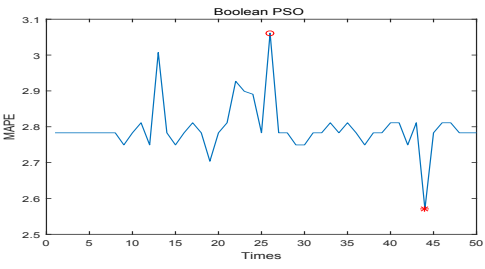

Fig. 4(b) MAPE comparison of Boolean PSO

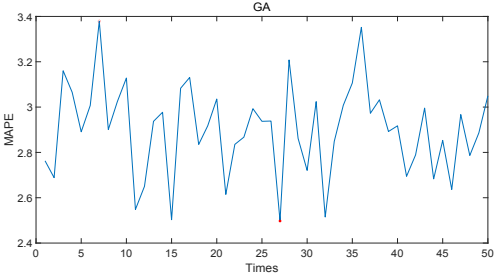

Fig.4(c) MAPE comparison of GA

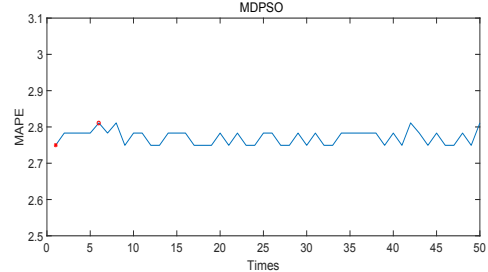

Fig.4(d) MAPE comparison of MDPSO

Fig. 4 MAPE comparison of four examined models in each modeling processing in China dataset

According to the Fig. 4(a-d), it is clear that DPSO has little fluctuation and remains stable at the value of 2.783, except for the special case of 2.749. MDPSO fluctuates slightly between

2.811 and 2.749. What DPSO and MDPSO have in common is that they consistently outperform Full input in all experiments, while the main difference between the two is how often they reach the minimum, specifically, DPSO reaches the minimum only once in 50 experiments, while MDPSO reaches the minimum in nearly half of the experiments. As we can see from the Fig. 4(b-c), BPSO has a significant fluctuation between 2.570 and 3.061, and BPSO is superior to Full input in the majority of experiments. GA fluctuates dramatically between 2.497 and 3.379, and GA is inferior to Full input in the majority of experiments.

As for comparisons across the five models, the experimental results presented in Table 1, Fig.3 and Fig.4 reveal some hints for common forecasting practice.

- The MAPE value of GA is different in each experiment in the two datasets. The possible reason is that GA method for feature selection can produce more diversified solutions, which makes the prediction results of SVR with different feature sets different. Meanwhile, GA is inferior to Full input in the majority of experiments in the two datasets, which implyes GA can't effectively select correct features set.

- MDPSO consistently outperforms Full input in 50 experiments in the two datasets according to MAPE measure, which indicates that MDPSO achieves more accurate forecasts than Full input. Thus, it is conceivable that MDPSO method for feature selection can improve the prediction accuracy of SVR modeling. Additionally, among the four feature selection strategies, MDPSO has relatively small fluctuation range in the two datasets, which shows that MDPSO method can stably select correct features.

### 5.2  Study 2: comparison with well-established counterparts

In this subsection, the results of SVR, BPNN, Holt-Winter and MDPSO-BPNN are presented for comparative analysis with the result of the proposed MDPSO-SVR strategy, which are shown in Table 2. Fig. 5 shows the fluctuation in the MAPE values of BPNN, Holt-Winter, MDPSO-SVR and MDPSO-BPNN models over 50 modeling experiments. Considering that the prediction results of the four models in each experiment are not unique, the corresponding prediction results when these models take the smallest MAPE value are shown in Fig. 6(a) and Fig. 6(b).

**Table 2.** Prediction accuracy measures of all the examined models in U.S and China datasets

| Dataset | Strategy | Accuracy | | | Elapsed Time |
|---|---|---|---|---|---|
| | | MAPE | RMSE | NRMSE | |
| U.S | SVR | 1.92 | 8.59 | 0.0263 | **0.11** |
| | BPNN | 4.07±0.91 | 17.13±4.05 | 0.0524±0.0124 | 13.51 |
| | Holt-Winter | 2.32±0.20 | 9.38±0.70 | 0.0287±0.0021 | 0.31 |
| | MDPSO-SVR | **1.82±0.04** | **8.28±0.22** | **0.0253±0.0006** | 4.01 |
| | MDPSO-BPNN | 4.69±1.48 | 20.89±7.91 | 0.0640±0.0242 | 651.51 |
| China | SVR | 2.85 | 256.86 | 0.0428 | **0.13** |
| | BPNN | 6.84±2.00 | 510.73±139 | 0.0852±0.0232 | 13.75 |
| | Holt-Winter | 3.88±0.15 | 291.85±7.01 | 0.0487±0.0012 | 0.57 |
| | MDPSO-SVR | **2.77±0.02** | **242.54±6.45** | **0.0404±0.0011** | 4.43 |
| | MDPSO-BPNN | 6.92±1.85 | 515.34±126 | 0.0859±0.0210 | 231.52 |

Note: For each row of table, the entry with smallest value is set in boldface.

As per the results presented in Table 2, it is not difficult to find that when considering the means of the three accuracies, the prediction performance of all examined models in the above two datasets is consistent as follows: among the three models of SVR, BPNN, and Holt-Winter, SVR consistently performed the best, followed by Holt-Winter, and finally BPNN. After introducing the MDPSO feature selection method, MDPSO-SVR consistently outperforms SVR and performs the best among all the models. While MDPSO-BPNN is consistently inferior to BPNN and performs the worst among all the models. When considering the standard deviations of the three accuracies, MDPSO-SVR is consistently the smallest in both datasets, followed by Holt-Winter. As for as the comparison of BPNN and MDPSO-BPNN, MDPSO-BPNN outperforms BPNN in the U.S case, while BPNN is better than MDPSO-BPNN in the China case. Additionally, the ranking with respect to the elapsed time from more to less are MDPSO-BPNN, then BPNN, then MDPSO-SVR, then Holt-Winter, and then SVR.

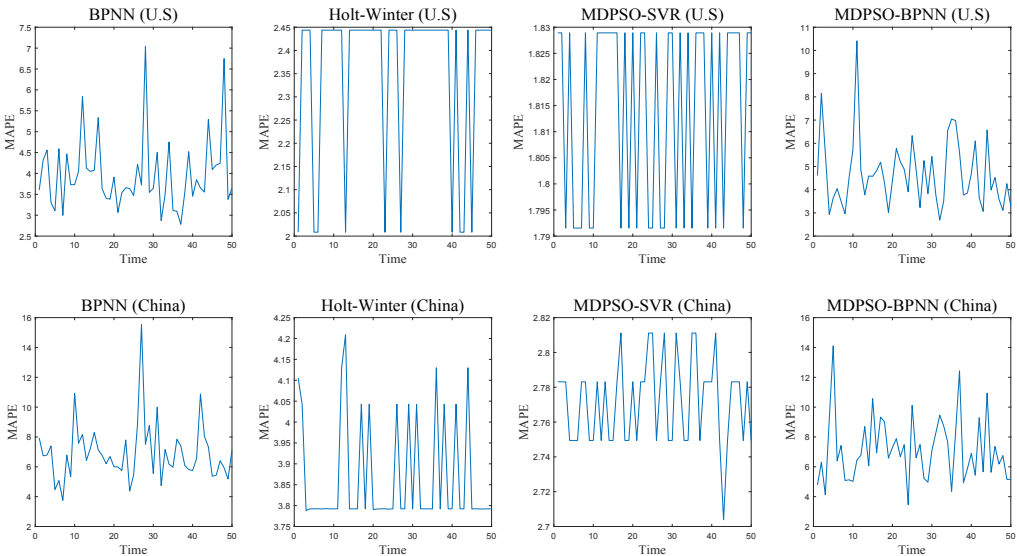

Fig. 5    MAPE comparison of different models in U.S and China datasets.

According to the results presented in the Fig. 5, one can deduce the following observation: In U.S case, MDPSO-SVR fluctuates slightly between 1.79 and 1.83, Holt-Winter fluctuates between 2.00 and 2.45, BPNN fluctuates sharply between 2.50 and 7.50, and MDPSO-BPNN fluctuates violently between 2.00 and 11.00. Obviously, the ranking with respect to fluctuation range from small to large is MDPSO-SVR, then Holt-Winter, then BPNN, and then MDPSO-BPNN. In China case, MDPSO-SVR fluctuates steadily between 2.70 and 2.82, Holt-Winter fluctuates between 3.75 and 4.25, BPNN and MDPSO-BPNN fluctuate significantly between

2.00 and 16.00. The ranking in terms of fluctuation range from small to large is MDPSO-SVR, then Holt-Winter, then BPNN and MDPSO-BPNN as a tie.

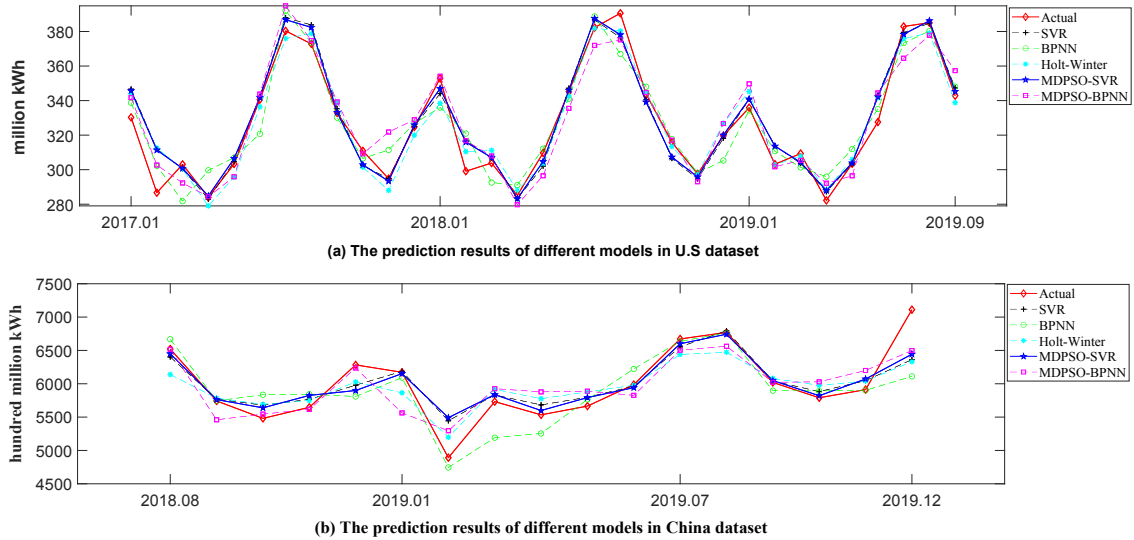

Fig. 6 The forecasted results of all the examined models in U.S and China datasets.

Overall, the experimental results presented in Table 2, Fig.5 and Fig.6 reveal some some common ground, as follows:

- Among the three models of SVR, BPNN, and Holt-Winter, SVR consistently performed the best, followed by Holt-Winter, and finally BPNN for each accuracy measure on the two datasets. There are two possible main reasons. One is that BPNN automatically adjusts many weight parameters during the running process, and once these weight parameters are set unreasonably, it will greatly weaken the prediction ability of BPNN. The other is that Holt-Winter usually predicts periodic series and extrapolates the initial periodic characteristics to predict, and once the periodic characteristics of the series change, it is difficult for the method to identify such changes and make improvements. Since SVR does not have the above problems, it has the best prediction performance among the three models.

- For the comparison of SVR and MDPSO-SVR, MDPSO-SVR consistently outperforms SVR for each accuracy measure on the two dataset. Meanwhile, MDPSO-SVR has relative small fluctuation range among all the examined models (except SVR). The possible reason is that during the SVR modeling process, MDPSO can select the correct feature set effectively and stably.

- As far as BPNN and MDPSO-BPNN are concerned, MDPSO-BPNN is consistently inferior to BPNN for each accuracy on the two datasets. The main reason could be that the result of each run of BPNN model is not unique, in this case, it is unreasonable to take the prediction performance of the BPNN model as an evaluation criterion to identify the correct input subset，that is why MDPSO method for feature selection cannot improve the prediction accuracy of BPNN modeling.

In general, the SVR model equipped with feature selection method using MDPSO algorithm can be a promised alternative for electricity consumption forecasting.

### 5.3  *Statistical test analysis*

In order to determine if there exists statistical significant difference among prediction performance of the above five models, analysis of variance (ANOVA) test procedures are performed for each performance measure. Specifically, SVR, BPNN, Holt-Winter, MDPSO-SVR and MDPSO-BPNN models are tested with the hold-out samples and the MAPE, RMSE and NRMSE are computed for 50 modeling process over U.S and China datasets. Supposed that $\mu_1, \mu_2, ..., \mu_5$ denote the means of MAPE of the above five models, then the null hypothesis of the ANOVA test is $\mu_1 = \mu_2 = \mu_3 = \mu_4 = \mu_5$. To judge whether the null hypothesis holds, the test statistic F is constructed as follows:

$$F = \frac{SSA/(k-1)}{SSE/(n-k)} \sim F(k-1, n-k)$$

where, SSA and SSE represent the between-group sum of squares and the within-group sum of squares, respectively. n and k denote the number of all observations and examined models, respectively. In this paper, n and k are equal to 250 and 5 respectively. When the statistic F is greater than the critical value $F_\alpha(k-1, n-k)$ of the given significance level $\alpha$, the null hypothesis is rejected, that is, the difference between the means is significant, and vice versa. The results of ANOVA test are shown in Table 3.

Table 3. The results of ANOVA test

| Dataset | Accuracy | Sum of squares | df | Mean of squares | Statistic F | p-Value |
|---|---|---|---|---|---|---|
| U.S | MAPE | SSA=340.207 | 4 | 85.052 | 142.992 | 0.000 |
| | | SSE=145.726 | 245 | 0.595 | | |
| | RMSE | SSA=5984.883 | 4 | 1496.221 | 81.836 | 0.000 |
| | | SSE=4479.379 | 245 | 18.283 | | |
| | NRMSE | SSA=0.057 | 4 | 0.014 | 82.114 | 0.000 |
| | | SSE=0.042 | 245 | 0.000 | | |
| China | MAPE | SSA=937.483 | 4 | 234.371 | 138.602 | 0.000 |
| | | SSE=414.286 | 245 | 1.691 | | |
| | RMSE | SSA=3854861.593 | 4 | 963715.398 | 100.202 | 0.000 |
| | | SSE=2356332.080 | 245 | 9617.682 | | |
| | NRMSE | SSA=0.113 | 4 | 0.028 | 88.747 | 0.000 |
| | | SSE=0.078 | 245 | 0.000 | | |

From Table 3, we can see that all ANOVA results are significant at the 0.05 level, which suggests that there is significant difference among the five examined models. To further identify the significantly different prediction models, Tukey's honesty significant difference (HSD) tests [40] are emplyed for multiple pair-wise comparisons. The results of the multiple comparison tests for two datasets are shown in Table 4. For each accuracy measure, we rank order the models from 1 (the best) to 5 (the worst).

Table 4. Multiple comparison results with ranked models for hold-out sample

| Dataset | Measure | Rank of Models | | | | | | | | |
|---|---|---|---|---|---|---|---|---|---|---|
| | | 1 | | 2 | | 3 | | 4 | | 5 |
| U.S | MAPE | MDPSO- | < | SVR | <* | Holt- | <* | BPN | <* | MDPSO-BPNN |
| | RMSE | MDPSO- | < | SVR | < | Holt- | <* | BPN | < | MDPSO-BPNN |
| | NRMSE | MDPSO- | < | SVR | < | Holt- | <* | BPN | < | MDPSO-BPNN |
| China | MAPE | MDPSO- | < | SVR | <* | Holt- | <* | BPN | < | MDPSO-BPNN |
| | RMSE | MDPSO- | < | SVR | < | Holt- | <* | BPN | < | MDPSO-BPNN |
| | NRMSE | MDPSO- | < | SVR | < | Holt- | <* | BPN | < | MDPSO-BPNN |

* The mean difference between the two adjacent methods is significant at the 0.05 level

As per the results presented in Table 4, one can make the following observations:

1) As far as the comparison of SVR, BPNN and Holt-Winter, SVR and Holt-Winter significantly outperform BPNN for all accuracy measures and datasets, and SVR significantly outperform Holt-Winter for the accuracy MAPE.

2) Considering the comparison MDPSO-SVR versus SVR, we can see that, although MDPSO-SVR consistently outperforms SVR for each accuracy measure and dataset, the difference in prediction performance between MDPSO-SVR and SVR is not significant at the 0.05 level.

3) Concerning the comparison of MDPSO-BPNN and BPNN, the difference in prediction performance is not significant at the 0.05 level, with a exception for the accuracy MAPE on U.S datasets. And the MDPSO-BPNN model performs the poorest at 95% statistical confifidence level in most cases.

**6. Conclusions**

Electricity consumption forecasting is an important issue in the energy planning in a country. The purpose of this study is to predict accurately monthly electricity consumption using the proposed MDPSO-SVR hybrid strategy. For this, we set two goals in the experimental study. One is to confirm the benefits of the MDPSO for feature selection in SVR forecasting model. The results show that MDPSO-SVR consistently outperforms SVR in 50 experiments in the two datasets according to MAPE measure, which indicates that MDPSO achieves more accurate forecasts than SVR. Thus, it is conceivable that MDPSO method for feature selection can improve the prediction accuracy of SVR modeling. In addition, among the SVR models with the DPSO, BPSO, GA and MDPSO four different feature selection methods, MDPSO-SVR has relative small fluctuation range in the two datasets, which shows that MDPSO method can select correct features relatively stably. The other is to examine the forecasting performance of the proposed MDPSO-SVR strategy. The results show that compared with SVR, BPNN, Holt-Winter and MDPSO-BPNN, MDPSO-SVR consistently performs best in the two datasets, which indicates the SVR model equipped with feature selection method using MDPSO algorithm can be a promised alternative for electricity consumption forecasting.

The limitation of this study is that the conclusions are drawn on the basis of univariate prediction model that merely use the lags of monthly electricity consumption as input variables. The reason for this is that this study aims to analysis the distinction of prediction performance under different feature selection strategies, and the results could be more reliable without considering exogenous variables. However, in terms of real-world scenarios, future efforts could be made to explore the improvement and development of hybrid modeling framework for electricity consumtion forecasting involving commonly used exogenous variables such as temperature, population, economic growth, and power facilities data.

**Acknowledgment**

This work was supported by Natural Science Foundation of China under Project Nos. 71871101, Jiangxi Principal Academic and Technical Leaders Program under project no.20194BCJ22015, and Science and Technology Research Projects of Jiangxi under project no. GJJ202905.